# Development of an interactive GUI using MATLAB for the detection of type and stage of Breast Tumor

Poulmi Banerjee
Department of Electronics & Communication Engg.
Regent Engineering and Research Foundation Group of Institutes
North 24 parganas, India
Email: banerjeepoulmi@gmail.com

Satadal Saha
Department of Electronics & Communication Engg.
MCKV Institute of Engineering
Howrah, India
Email: satadalsaha@mckvie.edu.in

*Abstract—* **Breast cancer is described as one of the most common types of cancer which has been diagnosed mainly in women. When compared in the ratio of male to female, it has been duly found that the prone of having breast cancer is more in females than males. Breast lumps are classified mainly into two groups namely: cancerous and non-cancerous. When we say that the lump in the breast is cancerous, it means that it can spread via lobules, ducts, areola, stroma to various organs of the body. On the other hand, non-cancerous breast lumps are less harmful but it should be monitored under proper diagnosis to avoid it being transformed to cancerous lump. To diagnose these breast lumps the method of mammogram, ultrasonic images and MRI images are undertaken. Also, for better diagnosis sometimes doctors recommend for biopsy and any unforeseen anomalies occurring there may give rise to inaccurate test report. To avoid these discrepancies, processing the mammogram images is considered to be one of the most reliable methods. In the proposed method MATLAB GUI is developed and some sample images of breast lumps are placed accordingly in the respective axes. With the help of sliders the actual breast lump image is compared with the already stored breast lump sample images and then accordingly the history of the breast lumps is generated in real time in the form of test report.**

## I. INTRODUCTION

Cancer or carcinoma is very common now-a-days. From Dr. Mary Ling's (Breast & General Surgeon) webpage, it is known that the part of the breast which produces milk is divided into 15-20 sections which are known as lobes. The ducts are the network of tiny tube through which milk travels which eventually exit the skin of the breast nipples. The areola is the dark area of the skin which surrounds the nipple portion. To provide the shape to the breast, the connective tissues and ligaments play an important role. The nerves in the breast provide the sensation. From Ann Pietrangelo's documentation in Healthline webpage, the early signs and symptoms of breast cancer are described as (a) development of a new lump in the region of breast or armpit, (b) feeling of swelling in some parts of the breast, (c) rashes and irritation occurs in the nipple area, (d) feeling of pain in any part of the breast, (e) changes is observed in the shape and size of the breast, (f) discharge of a fluid from the nipple apart from the breast milk.

A mass in Breast may occur due to abnormal cell division. There are basically two types of mass or breast tumors as follows:

*Benign tumors*: According to National Breast Cancer Foundation INC, when a tumor is diagnosed as non-spreading, doctors usually keep as it is rather than attempting to remove it. Though these tumors are not that alarming toward surrounding tissue, sometimes they may continue to spread towards the tissues and when any pain or other problem arises the tumor is removed, thus reliving the patient from pain or complications.

*Malignant tumors*: Malignant tumors are cancerous and may be dangerous because they enter and damage the surrounding tissues. According to National Breast Cancer Foundation INC, when a malignant tumor is found then the doctors advice to perform a biopsy to understand how severe the tumor is. In a human body, when the proto oncogene is activated and converted into oncogene then there is growth of cancerous cell. There are many reasons that are responsible for the conversion of proto oncogene to oncogene (heredity, gene mutation due to pollution etc).

## I. LITERATURE REVIEW

According to the research (G.C Wishart,et.al.,2010) done on the breast cancer detection in younger women it has been found that the process of mammography does not provide the desired result for the female having dense breast. The very recent improvements on this research work suggest that there might be a role play of this technology prior to performing needle core biopsy. Under this process the patients are scanned using digital infrared breast scan before the process of breast biopsy is performed. This research shows that the 106 biopsies performed in 100 women, 65 of the women were found to have malignant tumor and the rest 41 were found to be benign. Out of the four processes of biopsy the sentinel screening and sentinel neural network yielded an accuracy of 53% and 48% respectively while the analysis using the NoTouch software yielded an accuracy of 70% which was too close to the expert manual review which yielded an accuracy of 78%. The sensitivity and specificity detection using the technology of NoTouch BreastScan was found to be higher in the women of age group below 50 and combining mammogram and DIB with NoTouch BreastScan within this age group results in sensitivity of 89%.

The article on liquid biopsy by (Anna Babayan, et.al., 2018) suggest that the progress that has been made in sensitive analytical approaches has wide opened new arenas in the detection of cells such as circulating cell-free DNA that is being released by the tumors. The area of liquid biopsies are been now explored in clinical trials of the early detection of the breast cancer. In the very recent years the analysis process of the tumor cells and tumor derived products detectable in the blood and other body fluids coined by Pantel and Alix Panabieres as "liquid biopsy" has gained much importance.

According to (Ganesh N.Sharma, et.al., 2010), the breast cancer are of two types namely the non- invasive breast cancer and invasive breast cancer. The non-invasive breast cancer cells are only confined in the duct region and do not spread around the fatty and connective tissues of the breast. On the other hand, the invasive breast cancer cells have the tendency to break into the duct region along with the lobular walls of the breast thus spreading out to the surrounding fatty and connective tissues of the breast. From this research paper it is known that the most frequently occurring breast cancer is lobular carcinoma in situ and the ductal carcinoma in situ. Also, there is another type of breast cancer which is known as Invasive Lobular Carcinoma (ILC) and Invasive Ductal Carcinoma (IDC). Whenever cancer is detected, then and there a stage is being attached to it. Based on what stage the cancer is, the doctors determines the flow of treatment. The stages of the breast cancer are hereby divided into five categories: (a) *Stage 0*, where the doctors confirms of the tumor that is confined only to the milk duct and that it has not been spread to the surrounding tissues of the breast, (b) *Stage 1*, where the doctor confirms that the size of the tumor is less than 2 cm in diameter and that it has not reached outside the breast, (c) *Stage 2*, where the doctor confirms that the size of the tumor is between 2-5cm and that there may be the possibility that the tumor has spread to one of the three lymph nodes that are present in the armpit region or another possibility may arise that it has not reached to the surrounding of the breast, (d) *Stage 3*, where an inflammation in breast is observed and the size of the tumor ranges between 6-10cm and it is confirmed that the tumor has spread to the lymph nodes and the chest wall of the breast, (d) *Stage 4*, where the tumor regardless of its size spreads to every other organ.

According to (Elizabeth S.McDonald, et.al. 2016), the local regional staging while following up with the patient detected with the initial level of the breast cancer is sufficient through the physical examination, mammography etc. Sometimes the doctors recommend MRI for the patients of young age or may be someone for whom a mutation in gene is detected. Normally, when the cancer is detected at stage 1 or stage 2, the normal chest radiograph and the frequent lab blood test are sufficient for the diagnosis. When the cancer is detected at an advance stage, the National Comprehensive cancer network recommends CT scan of the chest, abdomen and pelvis along with the bone scan or sodium fluoride PET/CT scan.

According to (Jieun Koh, et. al. 2018) the AJCC committee removed the lobular carcinoma in situ from the in situ carcinoma (pTis) category and labeled it as the benign entity. The diseases namely the DCIS and Paget which occurs in the nipple are now not being in row with parenchymal carcinoma and hence are put into the categories of Tis (DCIS) and Tis (Paget disease). The invasive carcinoma is now being assigned to the category of T1-3 without the intervention of loco region which is basis on the invasive components size. For the estimation of the volume of the tumor, the maximum invasive tumor size is taken into account for the staging purpose. The $8^{th}$ edition as published by the AJCC committee holds a clarification that the largest tumor's maximum dimension be measured without any addition of the small microscopic satellite foci of the tumor. While the process of imaging it should be kept in mind that while measuring the largest tumor it should be included with the micro calcifications or architectural distortions size, which are again attached with the primary tumor. The tumor of category T1 is bounded within the dimension of 20mm or less and hence further sub divided into the categories of T1mi (with dimension ≤ 1mm), T1a (with dimension in the range 1mm-5mm), T1b (with dimension in the range 5mm-10mm) and T1c (with dimension in the range 10mm-20mm). Next, there is a category of T2 whose dimension lies in the range of 20mm - 50mm and T3 whose dimension lies in the range of > 50mm. The last category that is being assigned to the tumor is T4 when there is a possibility of invasion in the chest wall or the skin that is being caused by the cancerous cells of the breast.

According to the research (Virginia L, et.al, 2002), the first determination of sensitivity of the screening of mammography was detected for DCIS and invasive breast cancer. The sensitivity was defined based on the number of cases that resulted in a positive mammographic assessment during the process of screening divided by the total number of women present during that screening test. Next, on the basis of this, the percentage of the detected breast cancer of both screen and non-screen that were found to be DCIS and the rates of the same were done per 1000 mammograms divided among the four age groups of 40-49 years, 50-59 years, 60-69 years and 70-84 years. The denominator so obtained is then used to calculate the DCIS rate among all these age groups. From the individual BCSC sites the data so obtained shows that the percentages of cancers that have been reported to be DCIS and the rates of DCIS that have not been reported per 1000 mammograms. Based on the data obtained from pooled BCSC, it has been found out that 95% confident intervals were being used for the calculation of sensitivity, calculation of the percentage of cancers that have been reported as DCIS and for the rates of the reported DCIS per 1000 mammograms.

According to (Lilian C. Wang, et.al., 2013) the US examination was being conducted by using a 17-5 or 12-5 MHz linear array transducer. With the help of spring activated biopsy device, a 12 or 14 gauge needle was required. The process of imaging in a radial plane helps in the evaluation of the abnormalities in the ductal portion whereas in the antiradial plane this helps in analysis of margins. While setting non-calcified DCIS, the harmonic image processing proves helpful in identification of isoechoic lesions, which are very particularly found in the patients detected with fatty breast parenchyma. The increased breast density may be found to be obscure beneath the invasive component at mammography. The ultrasound scan performed on the breast shows that the calcification is not that evident as mammography. To guide into potential biopsy the ultrasound of calcification also be performed which has proved to be less costly than stereotactic biopsy and the most advantageous side of this test is that it does not involve ionizing radiation.

According to (Neha Sharma, et.al., 2016), the various techniques that are available for the segmentation of the images through mammography are image enhancement technique, histogram equalisation technique, watershed marker based technique, region growing technique, K-means clustering technique, Edge detection, transformation techniques etc. One of the softwares that can be used for segmentation of images is MATLAB. The most important issue in this software is to process the image that is of low quality to a image of high quality and to achieve this the intensity is enlarged producing a difference between the objects and its backgrounds thus yielding a high resolution representation of the breast tissues structure. After the enhancement of the contrast of the images the noise from the image is reduced by median filter in which a very efficient nonlinear filter removes the salt and pepper noise keeping the sharpness of the edge of image intake. This method of filtering assigns rank to the image pixels around the certain neighbourhood of a very specific pixel and the value is thus replaced by the median of neighbourhood. Along with this comes the process of segmentation which are again divided into two types namely histogram thresholding and slice technique.

According to (Basim Alhadidi et.al., 2007), as such no comprehensive imaging modality has been discovered for all the radiological applications, the ability for the computerization and the analysis of the medical images does provides a great means for assisting the physicians. The computer programs, processing methods to collect the data and information from the medical image scanners hence must be thoroughly developed for the enhancement of the critical clinical information. Along with the computer programs there must be something that would provide the diagnostic information which would be collected from the collected medical images and to achieve this there must be some computer algorithms which would help in enhancing the medical images. This is where the application based software MATLAB was introduced to detect breast cancer from the images collected from the mammogram.

According to (D.M.Garge et.al., 2009), the minimal cost wavelet based image processing technique using MATLAB can be one of the alternatives over the sophisticated computing platforms. In this method, the mammogram images are being collected from the web sources and then a noise is randomly added with a finite average value and then the noisy mammogram is then de-noised with the help of the threshold method. By using the MATLAB inbuilt function 'imfilter' along with wavelet and the sobel filter on the mammogram images, the further filtering process is then carried out. The calcified area of the breast in the mammogram images can thus be detected and then the matrix of the image is then carved out for the horizontal and vertical gradients.

According to (Yousif M.Y Abdallah et.al., 2018), the most powerful method to categorise breast normal tissues and the pathologies is the enhancement of images under the scanner of mammography. Using this digital image software there has come a great improvement in the mammographs as well as the illustration value also got a new dimension. The methods that were implemented in this paper are contrast improvement, noise lessening, texture scrutiny and portioning algorithm. It was also kept in mind that the mammographic images were kept in the highest quality. The main aim that was kept in mind while implementing this method was to get an augmented, noise free and honed intensified image.

From the above review it is clear that in spite of several research papers published to detect the malignancy of breast tumor, there is no GUI tool available for quantifying the parameters of a tumor, which may be helpful for deciding the malignancy of the tumor. The motivation behind the proposed work is addressing this issue.

## I. PROPOSED WORK

The proposed work focuses on the tumor type and the stages of the cancer, which will be determined on the basis of the diameter of the tumor. Also a test report is generated for the particular patient in a real time. Mammogram images are collected as the sampled data and MATLAB GUI platform is used for the purpose of image processing. Different types of technique like image erosion, image filtering are used for the detection of the breast tumor from the collected images for the purpose of noise reduction. Also, morphological dilation operation is applied for maintaining the proper shape of the tumor. An *imdistline* object is applied, which encapsulates a *Distance tool* consisting of an interactive *line* over an image paired with a *text label* that displays the distance between the line endpoints. This helps to find the diameter of the tumor. It is then converted from the pixel value into *cm* and/or *m* after calling another *sub GUI* and then according to the size of the tumor the results are obtained.

From the internet sample Images of the benign and malignant breast tumors are being collected along with the healthy one. Fig. 1 shows the flow diagram of the proposed work.

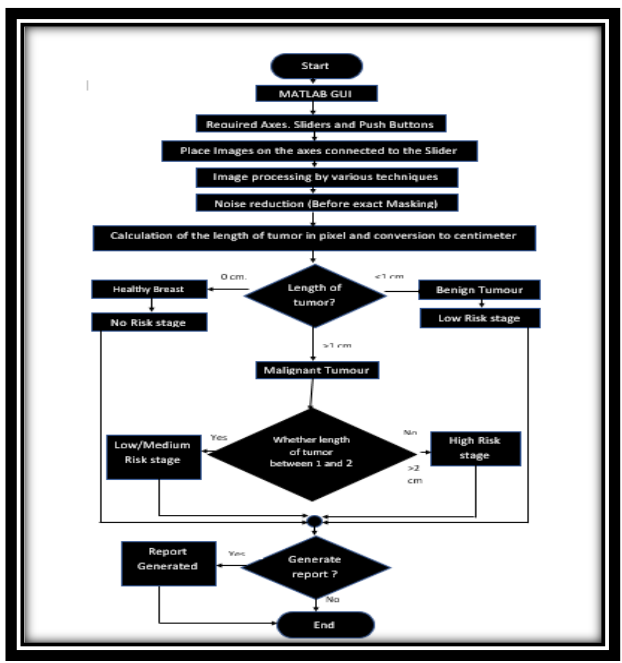

Fig.1 Flow diagram of the process.

Accordingly after the collection of images in MATLAB GUI, these images are placed on to the slider in the respective axes. The images collected from the internet were containing noise and to eradicate noise we use the technique of median filtering. It is a worldwide known popular method of preserving edges during the process of removal of noise. The main agenda to use this filter is scanning every single input data that would intercede with the entries that will come overall with the median function known as window method. Median filter is used to filter the image in two dimensional way. By using median filter function in each output pixel the value of the median is stored in a 3 by 3 neighborhood surrounding around the corresponding pixel of the image that is taken as input. Thresholding function is used for computing the global threshold parameter from its gray scale image format using a very popular method known as Otsu's method. This method prominently looks at each and every matching values corresponding to the threshold that lies between the foreground and background, thus calculating the variance that lies in between each of the two clusters. After this operation its selects the corresponding value against the variances that is the least. With the help of this method a threshold is being chosen that will minimize the interclass variance of black and white pixels that would be thresholded. Watershed transform method is being used to develop segment from the border regions into the distinct able object. Next, the technique of morphology is used, that processes particular images according to the shape from the data pool of image processing operation. There are mainly two types of morphological operation performed namely erosion and dilation. The process of erosion basically does image pixel shrinking and removes the boundaries of the object of the pixel. The process of dilation is just the opposite of erosion in which the region will be growing out from their set boundaries.

After performing all the above operations now for finding the overall diameter of the breast tumor the method of object encapsulation through a distance tool is used, that consists of an double headed square line which passes through diameter of the breast tumor. Through this method the diameter of the breast tumor in pixel format is fetched accordingly. After that on a new GUI the measured diameter is hence converted from pixel to centimeter or meter by using a multiplication factor. Then from the analysis of the breast tumor in cm we determine the size of the breast tumor followed by the assessment of risk factor the tumor is in. According to the clinical features of the breast tumor, following cases may result:

(a) If the tumor is less than 1 cm and greater than zero that means it has the possibility to be a benign tumor.

(b) If it's more than 1 cm that means it has a possibility to be malignant tumor.

(c) Otherwise it will be defined as healthy breast with no lump.

After the detection of the size of the breast tumor, the stages of the breast tumor is determined as follows.

(a) If the size of the tumor is within 2 cm, it will fall into low to medium risk stage of the breast cancer.

(b) If the size of the tumor is within 1 cm, it will fall into low risk stage and

(c) If the size of the tumor is greater than 2 cm, it will fall into high risk stage.

After all the analysis is being done, a detailed lab test report will be generated in real time.

I. RESULTS

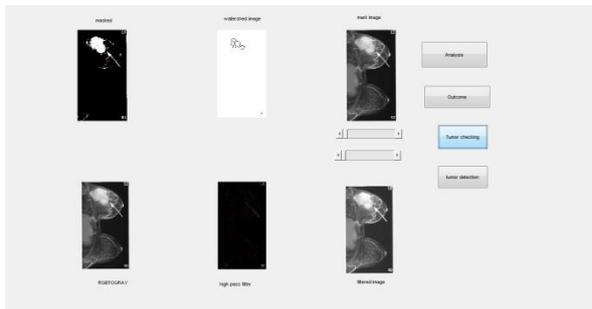

Fig. 2 Sample Images, filtered images, masked images are placed in various axes.

Pictures of the breast tumor which includes both malignant and benign. Alongside these two types of tumor healthy breast images are also being collected accordingly. The pictures are then being placed on the MATLAB GUI axes and navigated through the slider After setting of the images, through pushbuttons image analysis is being done basis on different methods like filtering, masking and watershed. (Fig: 2)

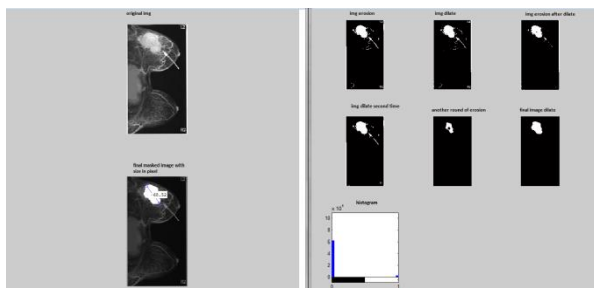

Fig. 3 Masking in a proper way and found out the tumor length

The exact shape of the breast tumor is being extracted out by passing through various stages of Dilation and Erosion. After that histogram of the related shape of the breast tumor is being generated. Immediately on the next window the original sample image is displayed along with the masked image of the breast tumor. Now the diameter of the breast tumor will be measured to perform analysis on the stage and the genre to which the tumor is belong to.(Fig: 3)

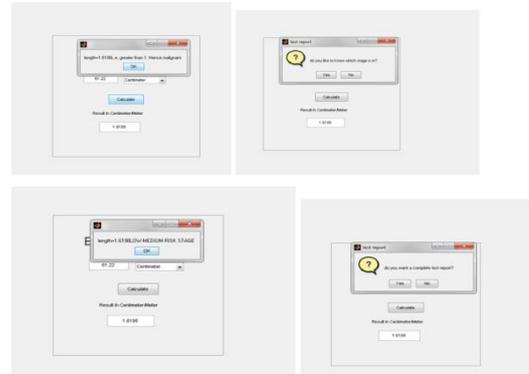

Fig. 4 Detection of length of tumor and identify types and stages of the tumor.

After analysis of the diameter of breast tumor in pixel, it is accordingly converted to centimeter. The diameter in centimeter will be displayed in a dialog box followed by the detailed report of the stage and the type of tumor. (Fig. 4).

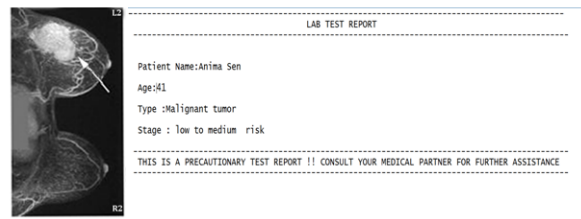

Fig. 5 Patient test report.

Another popup dialog box will be displayed asking whether to generate patient test report or not. If selected yes the patient test report will be generated in real time (Fig.5).

II. CONCLUSION

In a country like India where the resources to treat breast cancer is limited, there is a challenge to eradicate breast cancer fully here; but more efforts should be put into action so that it can lead to early detection and diagnosis of breast cancer. Effort should also be made in providing impromptu and cost effective treatment. So the process of image processing would be an effective tool in eradicating these barriers to some extent.

It can be concluded from various medical sources that the process of biopsy is very risky for cancer patients not only for breast cancer but also for chronic lymphatic leukemia where a bone marrow biopsy is done for beginning the treatment which is quite painful for every leukemia patient. Besides this there is also a delay in getting the reports which may further delay of the process of diagnosis as well as treatment.

Taking into consideration the above scenario, the image processing using MATLAB would provide an pain free diagnosis with proper result without any delay.

In this process after any screening test (MRI and MAMMOGRAM) is performed on the patient, the screening test image is sent for processing in MATLAB where the image is being segmented accordingly and with the help of intensity distribution graph a proper analysis of the stages of breast cancer is done. Apart from this during the process of scanning, the growth of the breast tumor is also being calculated which helps to know the nature of breast cancer.

This ensures a speedy preparation of the test report which surely would be helpful in early diagnosis of breast cancer.